\documentclass{article}
\usepackage{spconf,amsmath,graphicx}

\usepackage{enumitem}
\setlist{nosep, leftmargin=14pt}

\usepackage{mwe} 

\usepackage{subfig}
\usepackage{booktabs}
\usepackage{xcolor}

\usepackage{algorithm}
\usepackage{algorithmic}
\usepackage[T1]{fontenc}
\usepackage{hyperref}

\title{Scaling Neuroscience Research using Federated Learning}
\name{Author(s) Name(s)\thanks{Some author footnote.}}
\address{Author Affiliation(s)}
\twoauthors
 {Dimitris Stripelis, Jos\'{e} Luis Ambite}
	{Information Sciences Institute\\
	University of Southern California\\
	Marina del Rey, CA, USA}
 {Pradeep Lam, Paul Thompson}
	{Imaging Genetics Center\\
	USC Stevens Neuroimaging and Informatics Institute\\
	Marina del Rey, CA, USA}

\begin{document}
%
\maketitle
\begin{abstract}
The amount of biomedical data continues to grow rapidly. However, the ability to analyze these data is limited due to privacy and regulatory concerns. Machine learning approaches that require data to be copied to a single location are hampered by the challenges of data sharing. Federated Learning is a promising approach to learn a joint model over data silos. This architecture does not share any subject data across sites, only aggregated parameters, often in encrypted environments, thus satisfying privacy and regulatory requirements. Here, we describe our Federated Learning architecture and training policies. We demonstrate our approach on a brain age prediction model on structural MRI scans distributed across multiple sites with diverse amounts of data and subject (age) distributions. 
In these heterogeneous environments, our Semi-Synchronous protocol provides faster convergence.

\end{abstract}
\begin{keywords}
Deep Learning, Federated Learning
\end{keywords}
\section{Introduction}
\label{sec:Introduction}

Advances in computer technology, electronic medical records, and cost-efficient biomedical data acquisition, from sensors to genetic tests, allow healthcare organizations and biomedical research studies to collect increasing amounts of data. Analysis of these vast datasets using machine learning approaches promises novel discoveries. Unfortunately, privacy, security, and regulatory constraints make sharing datasets across studies or organizations extremely difficult, so that this promise is largely unfulfilled since joint analyses are limited. 

To address these challenges, Federated Learning  \cite{mcmahan2017communication,yang2019federated} has emerged as a novel privacy-preserving distributed machine learning paradigm that enables large-scale cross-institutional analysis without the need to move the data out of its original location. Federated Learning allows institutions to collaboratively train a machine learning model (e.g., a neural network) by aggregating the parameters (e.g., gradients) of local models trained on local data. Since subject data is not shared, and parameters can be protected through encryption, privacy concerns are ameliorated.  Even though Federated Learning was originally developed for mobile and edge devices, it is being increasingly applied in biomedical and healthcare domains \cite{lee2018privacy,sheller2018multi,silva2019federated,rieke2020future,silva2020fed}. 
We developed a Federated Learning architecture and training policies resilient to data and computational heterogeneity, where different sites may have different data amounts, target distributions, and computational capabilities \cite{stripelis2020semisynchronous,stripelis2020accelerating}, which are often characteristic of biomedical studies.

Brain age prediction from brain structural MRIs is a challenging biomedical task. The difference between the predicted and chronological brain age values is a phenotype related to aging and brain disease. Recent work \cite{cole2017predicting,jonsson2019brain} has shown that deep learning methods can accurately predict an individual's brain age. However, data scarcity limits the power of these methods, since privacy requirements make data sharing difficult. The
Federated Learning paradigm is a natural fit for these challenging learning environments. 

Here, we present our Federated Learning (FL) architecture and an empirical evaluation of brain age prediction under homogeneous, and heterogeneous environments with different amounts of data, and with data not independently and identically distributed (Non-IID) across sites \cite{yang2019federated,stripelis2020semisynchronous,stripelis2020accelerating,li2018federated}. We compare the effectiveness of the federated model to its centralized counterpart. We show that in heterogeneous environments, our communication-efficient Semi-Synchronous training policy \cite{stripelis2020semisynchronous} provides faster convergence.

\section{Related Work}
\label{sec:RelatedWork}

Federated Learning holds much promise in healthcare domains \cite{rieke2020future}. Silva et al. \cite{silva2020fed} present an open-source FL framework for healthcare, supporting different models and optimization methods. 
%
FL has been used for phenotype discovery \cite{liu2019two}, for patient representation learning \cite{kim2017federated}, and for identifying similar patients across institutions \cite{lee2018privacy}.

In biomedical imaging, Federated Learning has been applied to multiple tasks, including whole-brain segmentation of MRI T1 scans \cite{roy2019braintorrent}, brain tumor segmentation \cite{sheller2018multi,li2019privacy}, multi-site fMRI classification and identification of disease biomarkers \cite{li2020multi}, and for identification of brain structural relationships across diseases and clinical cohorts using (federated) dimensionality reduction from shape features \cite{silva2019federated}. COINSTAC \cite{plis2016} provides a privacy-preserving distributed data processing framework for brain imaging. 

Depending on the requirements and computational characteristics of the federated learning environment, the participating sites (learners) can be organized under different topologies \cite{yang2019federated,rieke2020future}. In a star topology (Figure~\ref{fig:FederatedLearningEnvironmentArchitecture}), the learners communicate through a head server that is responsible for coordinating the federation training (e.g., our work and \cite{sheller2018multi,li2019privacy,li2020multi}). In a peer-to-peer topology \cite{roy2019braintorrent}, learners can communicate directly with each other without a distinguished coordinator. 


Most current FL approaches  \cite{mcmahan2017communication,li2018federated,bonawitz2019towards} compute the global model through a synchronous communication protocol. However, in heterogeneous environments, stragglers may slow down convergence. Our Semi-Synchronous protocol \cite{stripelis2020semisynchronous} counters this inefficiency by assigning more local computation to the underutilized learners to accelerate convergence.

%

Preserving data privacy is critical in Federated Learning environments. Common methods for privacy protection are differential privacy \cite{abadi2016deep} and homomorphic encryption \cite{zhang2020batchcrypt}. For example, the federated learning system for brain tumor segmentation in \cite{li2019privacy} used differential privacy techniques. 
%
We are developing a homomorphic encryption approach in our architecture, but it is out of the scope of this paper. 

Finally, our deep learning model for estimating brain age from structural MRI scans is closely related (albeit different) to \cite{cole2017predicting,jonsson2019brain,PENG2020101871}.
However, we use Federated Learning to learn the joint model, as opposed to using majority voting and linear regression data blending to combine CNN brain age predictions from different data sources as in \cite{jonsson2019brain}.

\section{Federated Learning}

\begin{figure}[htpb]
  \centering
  \includegraphics[width=\linewidth]{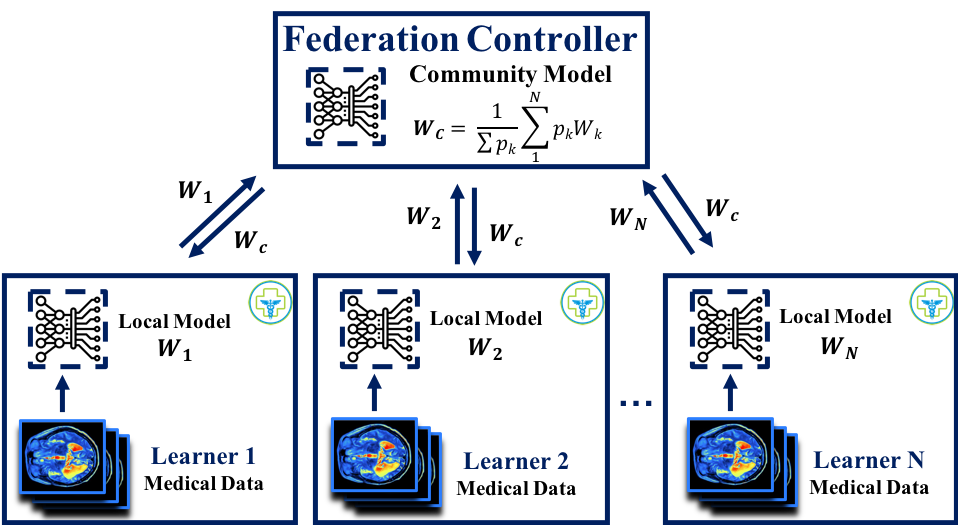}
  \captionsetup{justification=centering}
  \caption{Federated Learning Architecture}
  \label{fig:FederatedLearningEnvironmentArchitecture}
\end{figure}

Federated Learning operates over sites that do not share data, so the joint model is obtained through parameter sharing and mixing \cite{yang2019federated,rieke2020future}. 
Figure \ref{fig:FederatedLearningEnvironmentArchitecture} shows the typical Federated Learning architecture. Here, we consider that a single neural network, known to all the sites, is being optimized. 
Each site (learner) trains the neural network on its own local private dataset and shares only the locally-learned parameters with a head server, the Federation controller, which is responsible for aggregating the local parameters to generate a community neural network, which is in turn sent back to the learners. 

\textbf{Federated Optimization.} The primary goal in Federated Learning is to jointly learn a global/community model across a federation of N learners, with non-co-located data, by optimizing the global objective $f(w)$:
\begin{equation}\label{eq:FederatedFunction}
w^*=\underset{w}{\mathrm{argmin}} f(w) \:\: \text{with} \:\: f(w)=\sum_{k=1}^{N}\frac{p_k}{\sum p_k}F_k(w)
\end{equation}
%
\noindent where $p_k$ is the contribution value of a learner $k$ to the federation and $F_k(w)$ its local objective function. 
%
%
The contribution value $p_k$ can be static, or dynamically defined at run time~\cite{stripelis2020accelerating}. In much recent work \cite{mcmahan2017communication, bonawitz2019towards}, the learners are weighted based on the number of local training examples ($p_k=|D_{k}^T|$), since this is a good proxy for the value of a local model, but other methods that directly measure performance are possible~\cite{stripelis2020accelerating}.

\textbf{Synchronous Training.} In the original Federated Learning algorithm  \cite{mcmahan2017communication,bonawitz2019towards}, every learner performs a predefined number of local updates (batches or epochs) before reaching a synchronization point where it shares its local model with the federation. This computational approach, which we refer to as Synchronous Federated Average (\textit{SyncFedAvg}), has been extensively explored \cite{mcmahan2017communication,li2018federated,bonawitz2019towards}.

\textbf{Semi-Synchronous Training.} We introduced a Semi-Synchronous training protocol (\textit{SemiSync}) \cite{stripelis2020semisynchronous} where each learner trains for a given amount of time before synchronization. Each learner processes a variable number of data batches between synchronization points depending on its computational power and amount of data. 
SemiSync parameterizes the synchronization period based on the time that it takes for the slowest learner in the federation to perform a single epoch. The number of local updates (batches) $\mathcal{B}_k$ a learner $k$ performs between synchronization points is computed as:
\begin{equation}\label{eq:SemiSynchronousScheduling}
    \begin{gathered}
        t_{max}(\lambda) = \lambda * \max \limits_{k \in N} {\{\frac{|D_k^T|}{\beta_k} * t_{\beta_k}\}}, \lambda, \beta_k,t_{\beta_k} >0 \\
        \mathcal{B}_k = \dfrac{t_{max}}{t_{\beta_{k}}}, \quad \forall k \in N
    \end{gathered}    
\end{equation}
\noindent where $|D_k^T|$ is the number of local training examples, $\beta_k$ is the learner's local batch size defined at global model initialization and $t_{\beta_k}$ is the time it takes to perform a local batch (i.e., an update). The hyperparameter $\lambda$ controls the communication frequency of the learners by adjusting the number of local updates per learner based on the time it takes for the slowest learner to perform $\lambda$ local epochs.

This training policy is particularly effective in federated learning settings where learners have homogeneous computational power, but heterogeneous amounts of data, as well as in settings where learners have heterogeneous computational power and/or heterogeneous amounts of data \cite{stripelis2020semisynchronous}.

\section{Federated Brain Age Prediction}
\label{sec:BrainAgeModel}

The learning task we investigate here is brain age prediction. Deep 3D convolutional regression networks have been used for brain age prediction \cite{cole2017predicting,jonsson2019brain}. These networks extend the VGG and ResNet architectures to 3D images by replacing 2D convolution/maxpool operations with their 3D counterparts.

\begin{figure}[htb]
    \centering
    \includegraphics[width=\linewidth]{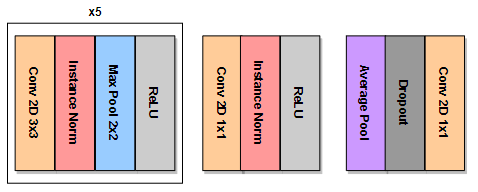}
    \caption{BrainAgeCNN}
    \label{fig:brainage_cnn}
\end{figure}

\textbf{Neural Architecture.} 
Figure~\ref{fig:brainage_cnn} shows the convolutional encoding network we trained for the brain age prediction task. The model architecture is similar to that in \cite{PENG2020101871} with the main difference being the replacement of the batch normalization (BatchNorm) layer with an instance normalization (InstanceNorm) layer. Collectively, the network consists of seven blocks, with the first five composed of a 3x3x3 3D convolutional layer (stride=1, padding=1), followed by an instance norm, a 2x2x2 max-pool (stride=2), with ReLU activation functions. The number of filters in the first block is 32 (and doubles until 256) with both layers 4 and 5 having 256 filters. The sixth block contains a 1x1x1 3D convolutional layer (stride=1, filters=64), followed by an instance norm and ReLU activation. The final, seventh, block contains an average pooling layer, a dropout layer (set to p=0.5 during training), and a 1x1x1 3D convolutional layer (stride=1). To train the model we used Mean Squared Error as loss function and Vanilla SGD as the network's optimizer. 

\vspace{-1mm} 
\vspace{-1mm}  


\textbf{Federated Model.} During federated training all learners train on their local data using the same neural architecture and hyperparameters (e.g., learning rate, batch size). Once a learner finishes its local training, it sends its local model parameters to the controller. 


\section{Experiments}
\label{sec:Experiments}

Our goal is to apply Federated Learning to hospital consortia, and to large research studies like Enigma (enigma.ini.usc.edu). As an initial step in a controlled environment, we analyzed brain MRI data from the UK Biobank \cite{miller2016multimodal}, a large epidemiological study of 500,000 people residing in the United Kingdom, some with neuroimaging.
%
We explored several heterogeneous federated learning scenarios with different data distributions and amounts of data per learner and evaluated the performance and convergence rate of the federation. 

\textbf{NeuroImaging Data.} From the original UKBB dataset of 16,356 individuals with neuroimaging, we selected 10,446 who had no indication of neurological pathology, and no psychiatric diagnosis as defined by the ICD-10 criteria. The age range was 45-81 years (mean: 62.64; SD: 7.41; 47\% women, 53\% men). All image scans were evaluated with a manual quality control procedure, where scans with severe artifacts were discarded. The remaining scans were processed using a standard preprocessing pipeline with non-parametric intensity normalization for bias field correction1 and brain extraction using FreeSurfer and linear registration to a (2 mm)$^3$ UKBB minimum deformation template using FSL FLIRT. The final dimension of the registered images was 91x109x91. 
The 10,446 records were split into 8356 for train and 2090 for test.
%

\vspace{-5mm} 

\begin{figure}[htpb]
  \captionsetup{justification=centering}
  \subfloat[Uniform \& IID \newline Age Buckets]{
    \includegraphics[width=0.32\linewidth]{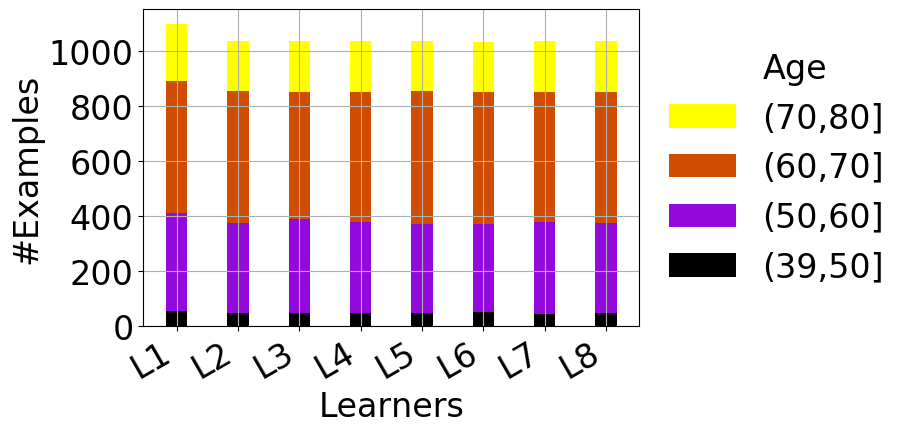}
    \label{subfig:UKBB_AgeBuckets_Uniform_IID}
  }
    \subfloat[Uniform \& Non-IID Age Buckets]{
    \includegraphics[width=0.32\linewidth]{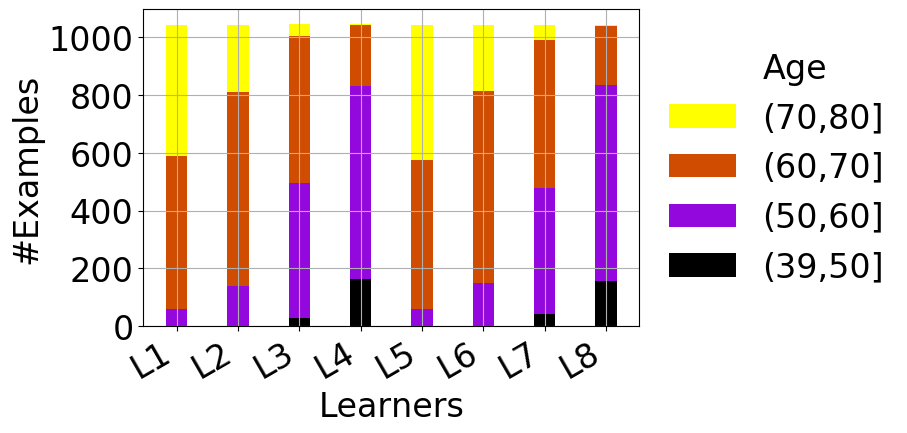}
    \label{subfig:UKBB_AgeBuckets_Uniform_NonIID}
  }
    \subfloat[Skewed \& Non-IID Age Buckets]{
    \includegraphics[width=0.32\linewidth]{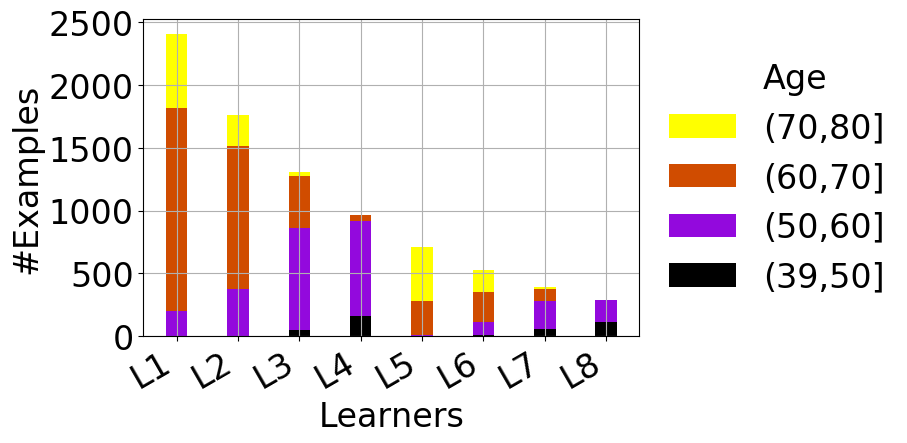}
    \label{subfig:UKBB_AgeBuckets_Skewed_NonIID}
  }
  
  \subfloat[Uniform \& IID \newline Age Distribution]{
  \centering\includegraphics[width=0.32\linewidth]{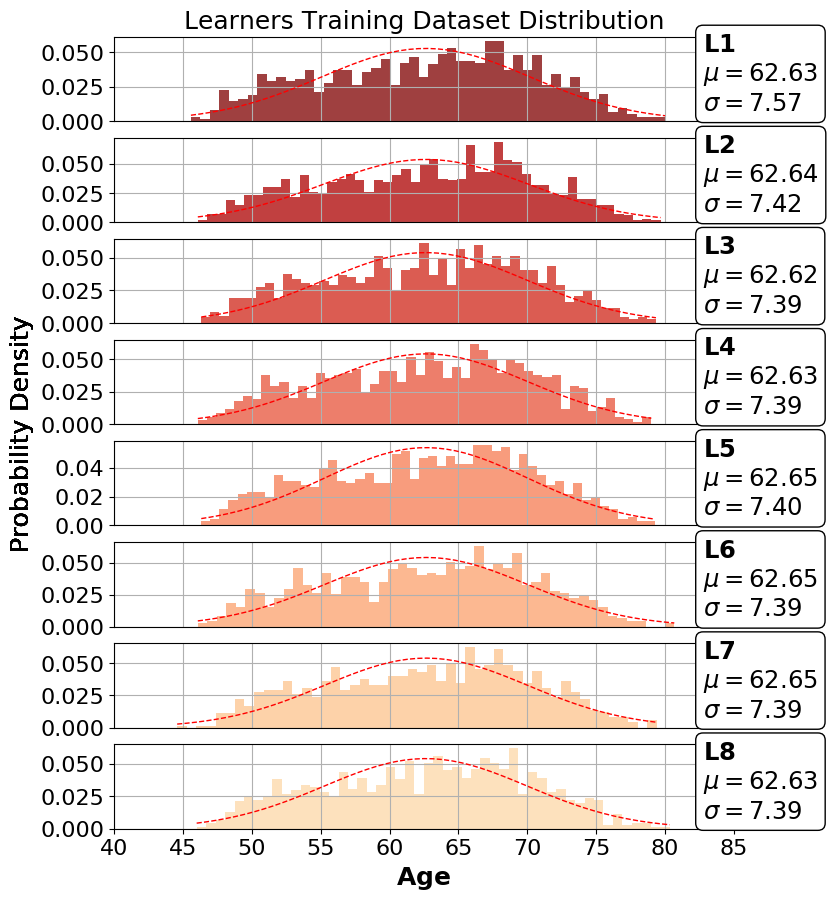}
  \label{subfig:UKBB_AgeDistribution_Uniform_IID}
  }
  \subfloat[Uniform \& Non-IID Age Distribution]{
  \centering\includegraphics[width=0.32\linewidth]{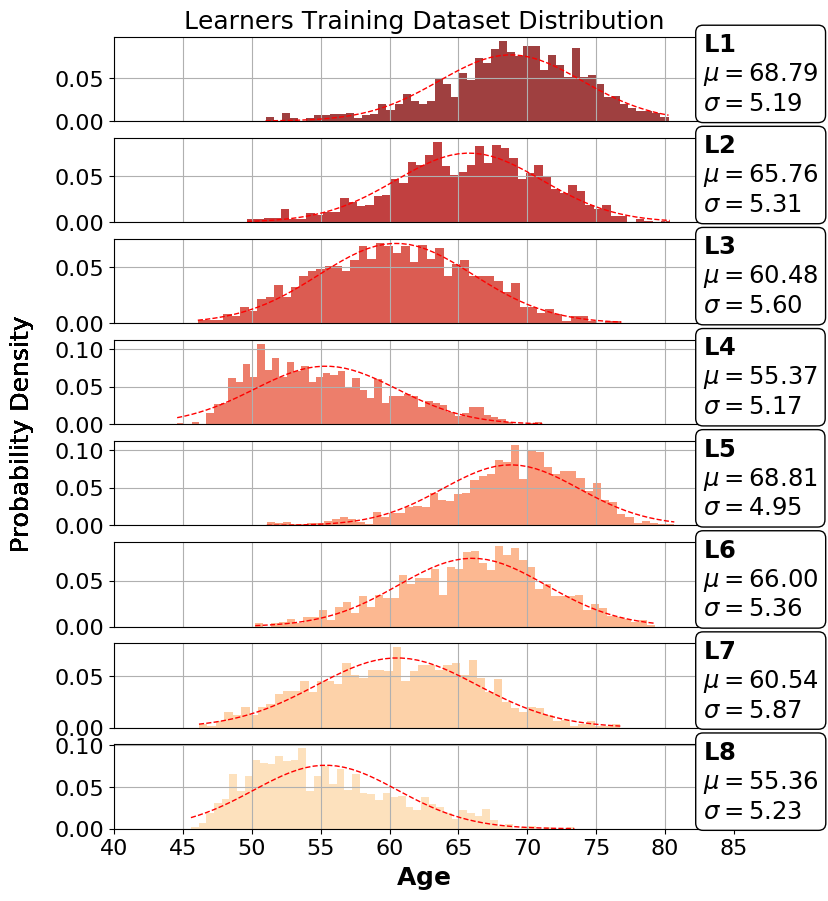}
  \label{subfig:UKBB_AgeDistribution_Uniform_NonIID}
  }
  \subfloat[Skewed \& Non-IID Age Distribution]{
  \centering\includegraphics[width=0.32\linewidth]{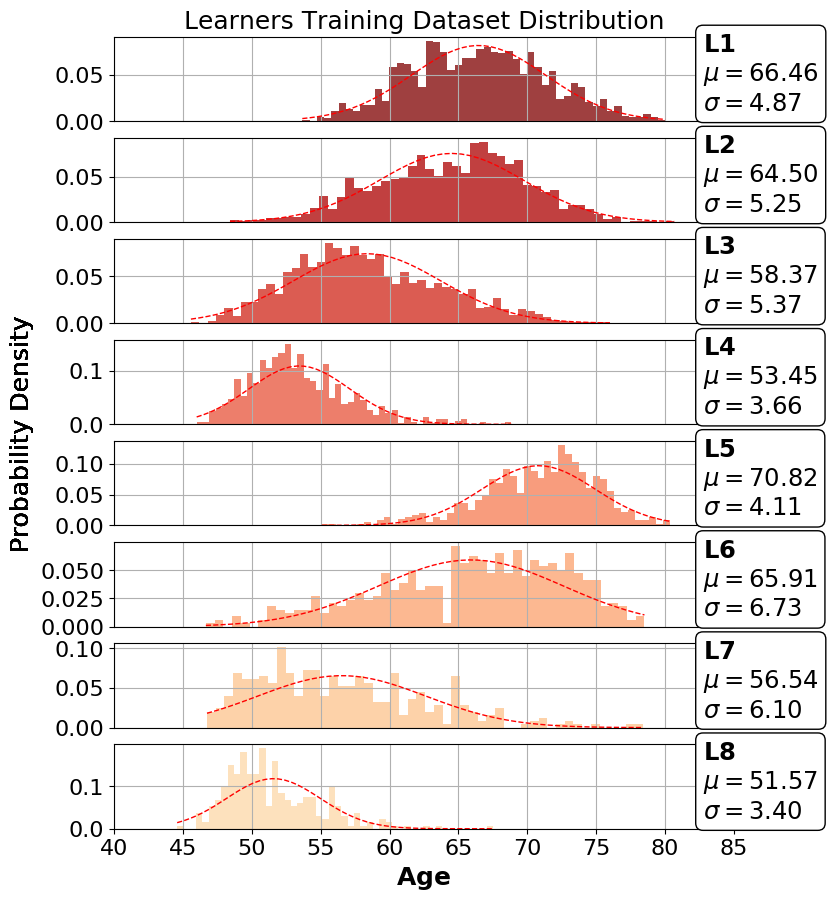}
  \label{subfig:UKBB_AgeDistribution_Skewed_NonIID}
  }

  \captionsetup{justification=centering}
  \caption{UKBB Federation Data Distributions}
  \label{fig:UKBB_Federation_Age_Probability_Distribution}
  
\end{figure}


\textbf{Data Distributions.} We define several challenging learning environments by partitioning the centralized UKBB neuroimaging training dataset (8356 records) across a federation of 8 learners.%
\footnote{Available at: \url{https://dataverse.harvard.edu/dataset.xhtml?persistentId=doi:10.7910/DVN/2RKAQP}}
 Every environment was evaluated on the same test dataset (2090 records) and the learners used their allocated records for training.
As shown in Figure~\ref{fig:UKBB_Federation_Age_Probability_Distribution}: panels (a, b, c) show the amount of data (and age buckets), and the corresponding (d, e, f) show the detailed age distribution of each of the 8 learners. 
Figures~\ref{fig:UKBB_Federation_Age_Probability_Distribution}(a,d) show a uniform (same amount of data per learner) and IID (all ages) distribution. 
Figures~\ref{fig:UKBB_Federation_Age_Probability_Distribution}(b,e) show a uniform, but non-IID (subset of ages) distribution. 
Figures~\ref{fig:UKBB_Federation_Age_Probability_Distribution}(c,f) show a skewed (different amount of data per learner) and non-IID distribution.

\textbf{Training Environment.} We established a federation of 8 learners by assigning one learner to each GPU of a server with 8~GeForce GTX 1080 Ti graphics cards (10~GB RAM each), 40~Intel(R) Xeon(R) CPU E5-2630 v4 @ 2.20GHz, and 128GB DDR4 RAM. All learners trained on the same CNN model (Fig.~\ref{fig:brainage_cnn}). For both centralized and federated models, we used Vanilla SGD with a learning rate of $5\text{x}10^{-5}$ and a batch size ($\beta_k$) of 1. 
For SyncFedAvg, each learner runs 4 local epochs in all distributions. 
For SemiSyncFedAvg, the time per batch ($t_{\beta_{k}}$) for every learner is 120ms, the maximum time of a single epoch across all learners is 280secs (learner~1 that holds the largest partition: $\sim$2,400 examples), thus for $\lambda=4$ the maximum time ($t_{max}$) is 1,120secs (cf. Eq. \ref{eq:SemiSynchronousScheduling}). Finally, the assigned number of local updates ($\mathcal{B}_k$) per learner is close to 9300. For all experiments, the random seed was 1990.

\begin{figure}[htpb]
  \captionsetup{justification=centering}
  \centering
  \subfloat[Wall-Clock Time]{
  \centering\includegraphics[width=0.8\linewidth]{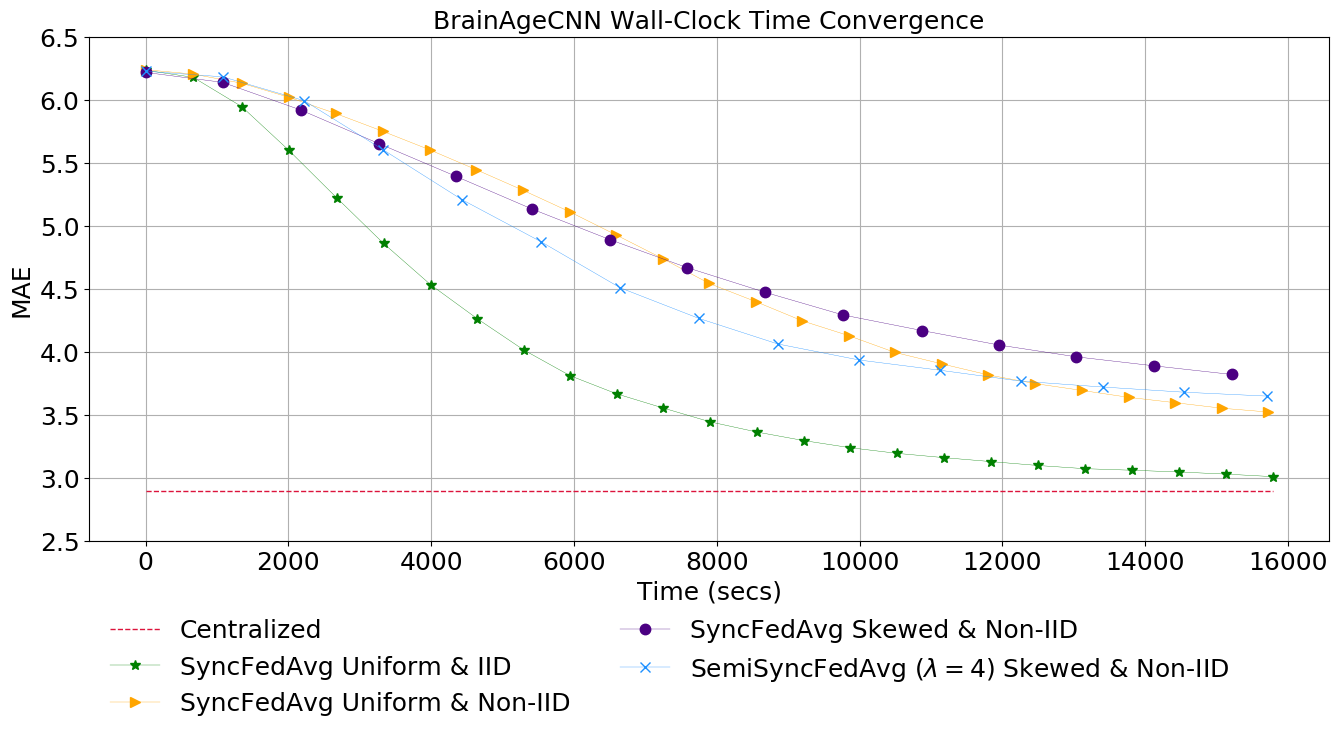}
  \label{subfig:UKBB_PoliciesConvergence_WallClockTime}
  }
  
  \subfloat[Federation Rounds]{
\includegraphics[width=0.8\linewidth]{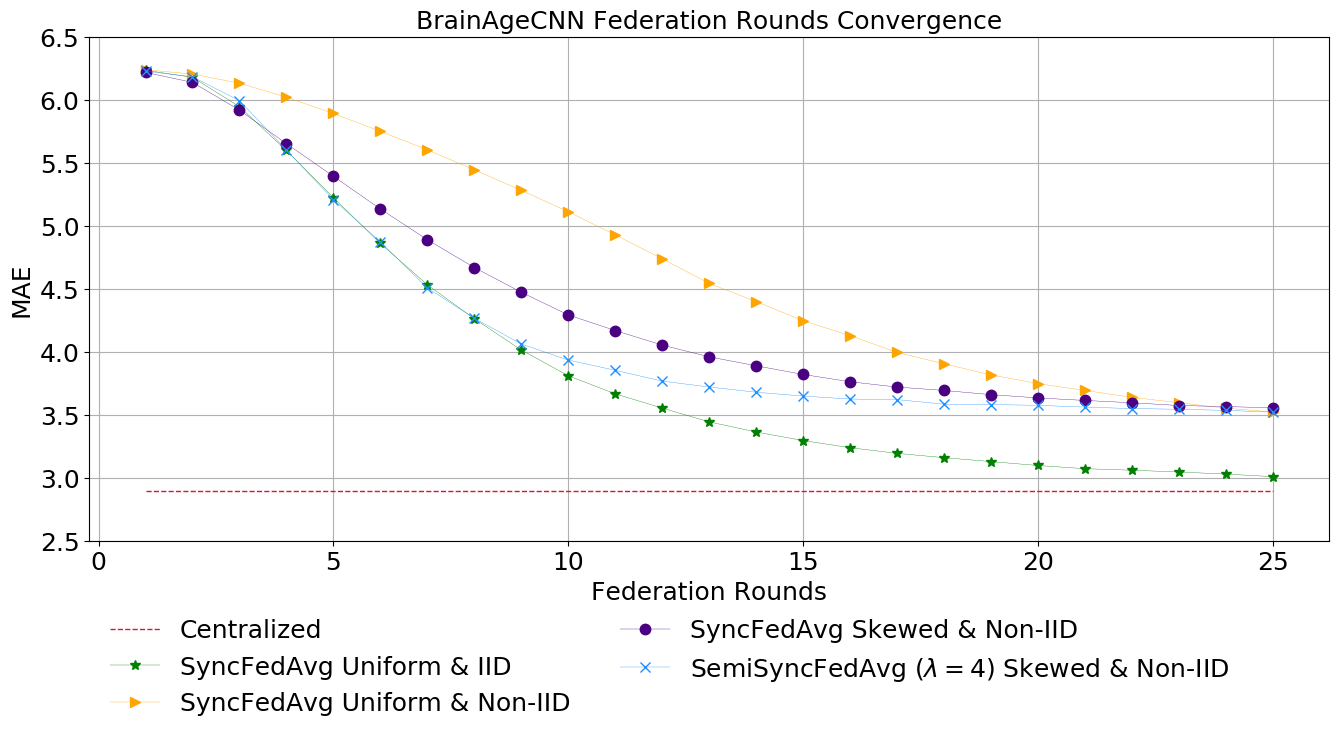}
  \label{subfig:UKBB_PoliciesConvergence_FederationRounds}
  }
  
  \captionsetup{justification=centering}
  \caption{UKBB Brain Age Federated Policies Performance}
  \label{fig:UKBB_FederatedPoliciesConvergence}
\end{figure}


\textbf{Evaluation.} 
We compare the performance of brain age prediction in settings with uniform and skewed data sizes, and with IID and non-IID age distributions. Since we consider a homogeneous computational environment with 8 identical learners (GPUs) and the computational cost of each learner only depends on the amount of data, for uniform data sizes we only show results for the Synchronous Federated Average policy (SemiSync performs the same). For skewed data sizes, we show both synchronous and semi-synchronous policies. 

Figure \ref{subfig:UKBB_PoliciesConvergence_WallClockTime} shows the performance of the training policies over the different environments in terms of elapsed (wall-clock) time (i.e., the 8 learners running in parallel). 
For the Uniform and IID setting, the federation reaches a Mean Absolute Error (MAE) value that is very close to the one achieved by the centralized model. More challenging Non-IID settings lower the performance, with a final error of 0.5 years over the centralized model (cf. Table~\ref{table:UKBB_Evaluation}). 
%
%
The Semi-Synchronous FedAvg policy has a faster convergence and slightly better final performance than Synchronous FedAvg. Even though the computational power of each learner is the same (identical GPUs), in the skewed data setting for the SyncFedAvg policy the learners with the smaller amounts of data remain idle until the learner with the most data finishes its allocated epochs. In contrast, the SemiSync policy continuously processes batches without any idle time. This additional computation results in the improved convergence in this setting. 

Figure~\ref{subfig:UKBB_PoliciesConvergence_FederationRounds} shows the performance in terms of federation rounds, which is a proxy for the communication cost of the policy.%
\footnote{Note that the wall-clock time for each federation round depends on the data distribution. To process the 25 federation rounds in Figure~\ref{subfig:UKBB_PoliciesConvergence_FederationRounds}, SyncFedAvg takes 15,482 seconds in the Uniform and IID setting, 15,777 seconds in the Uniform and Non-IID setting, and 22,713 seconds in the Skewed and Non-IID setting. For 25 federations rounds in the Skewed and Non-IID setting SemiSyncFedAvg takes 23,048 seconds. 
}
At the end of the allocated epochs for SyncFedAvg, or $\lambda$ time for SemiSyncFedAvg, all learners share their models with the Federation Controller, which then computes the community model and sends it back to the learners. Thus, the number of models exchanged through the network is twice the number of federation rounds times the number of learners. As before, SemiSyncFedAvg shows faster convergence in terms of communication cost than SyncFedAvg. 

\vspace{-3mm} 

\begin{table}[htpb]
\noindent
\tiny
  \begin{center}
    \begin{tabular}{@{}llccccc@{}}
      & & \textbf{MSE} & \textbf{RMSE} & \textbf{MAE} & \textbf{Corr}\\
      \toprule
      \multicolumn{2}{l}{\textbf{Centralized Model}} & 12.885 $\pm$ 0.021 & 3.589 $\pm$ 0.003 & 2.895 $\pm$ 0.006 & 0.881\\
      \toprule
      \multicolumn{2}{l}{\textbf{Federated Model}} &  &  &  & \\
      \textbf{Data Distribution} & \textbf{Policy} & & & &\\
      \cmidrule(r){1-2}
      \textbf{Uniform \& IID} & SyncFedAvg & 13.749 $\pm$ 0.138 & 3.707 $\pm$ 0.018 & 2.995 $\pm$ 0.018 & 0.875 \\
      \midrule
      \textbf{Uniform \& Non-IID} & SyncFedAvg & 19.853 $\pm$ 1.347 & 4.453 $\pm$ 0.151  & 3.625 $\pm$ 0.135 & 0.861\\
      \midrule
      \textbf{Skewed \& Non-IID} & SyncFedAvg & 19.148 $\pm$ 0.086 & 4.376 $\pm$ 0.009 & 3.553 $\pm$ 0.003 & 0.851\\
      & SemiSync($\lambda$=4) & 18.491 $\pm$ 0.122 & 4.311 $\pm$ 0.015 & 3.505 $\pm$ 0.008 & 0.864\\
      \bottomrule
    \end{tabular}
  \end{center}
  \captionsetup{justification=centering}
  \vspace{-1mm} 
  \caption{UKBB Evaluation. Mean and std values for 3 runs.}
  \label{table:UKBB_Evaluation}
\end{table}

\vspace{-3mm} 

\section{Conclusion}
\label{sec:Conclusion}

We have demonstrated the effectiveness of the Federated Learning paradigm in the neuroimaging domain by collaboratively learning a global model for brain age prediction. We empirically evaluated the convergence of the federated model in statistically heterogeneous learning environments. 
Our immediate future work includes investigating additional learning tasks, such as disease prediction, 
and incorporating homomorphic encryption in the architecture
%
We simulated the non-IID case by subsampling UKB data, but we plan to examine the more realistic case where learners receive data from different scanning protocols and cohorts, which would exhibit natural acquisition and population differences. This will better show the different relative performance of the different training policies.
Finally, we plan to explore federated transfer learning in neuroimaging.


\section{Acknowledgments}
\label{sec:Acknowledgements}
This research was supported in part by the Defense Advanced Research Projects Activity (DARPA) under contract HR0011\-2090104, and in part by the National Institutes of Health (NIH) under grants U01AG068057 and RF1AG051710.  The views and conclusions contained herein are those of the authors and should not be interpreted as necessarily representing the official policies or endorsements, either expressed or implied, of DARPA, NIH, or the U.S. Government.

\section{Compliance with Ethical Standards}
This is a study of previously collected, anonymized, de-identified data, available in a public repository. 
Data access approved by UK Biobank under Application Number 11559.





\bibliographystyle{IEEEbib}

\end{document}